\pdfoutput=1

\documentclass[11pt]{article}

\usepackage[preprint]{acl}

\usepackage{times}
\usepackage{latexsym}

\usepackage[T1]{fontenc}

\usepackage[utf8]{inputenc}

\usepackage{microtype}

\usepackage{inconsolata}

\usepackage{graphicx}

\usepackage{hyperref}

\usepackage{amsmath}
\usepackage{amssymb}
\usepackage{mathtools}
\usepackage{amsthm}

\theoremstyle{plain}
\newtheorem{theorem}{Theorem}[section]

\theoremstyle{definition}
\newtheorem{definition}[theorem]{Definition}

\theoremstyle{remark}

\usepackage[textsize=tiny]{todonotes}

\usepackage{pgfplots}
\usepackage{float}
\usepackage{multirow}
\usepackage{placeins}

\title{CARVQ: \underline{C}orrective \underline{A}daptor with Group \underline{R}esidual \underline{V}ector \underline{Q}uantization for LLM Embedding Compression}

\author{
 \textbf{Dayin Gou\textsuperscript{*}},
 \textbf{Sanghyun Byun\textsuperscript{*}},
 \textbf{Nilesh Malpeddi},
 \textbf{Gabrielle De Micheli},
\\
 \textbf{Prathamesh Vaste},
 \textbf{Jacob Song},
 \textbf{Woo Seong Chung\textsuperscript{\textdagger}}
\\
\\
LG Electronics USA
}

\begin{document}
\maketitle
\def\thefootnote{*}\footnotetext{Equal Contribution}
\def\thefootnote{\textdagger}\footnotetext{Corresponding Author}

\begin{abstract}



Large Language Models (LLMs) typically rely on a large number of parameters for token embedding, leading to substantial storage requirements and memory footprints. 
In particular, LLMs deployed on edge devices are memory-bound, and reducing the memory footprint by compressing the embedding layer not only frees up the memory bandwidth but also speeds up inference. 
%
To address this, we introduce CARVQ, a post-training novel Corrective Adaptor combined with group Residual Vector Quantization. CARVQ relies on the composition of both linear and non-linear maps and mimics the original model embedding to compress to approximately 1.6 bits without requiring specialized hardware to support lower-bit storage.
%
%
We test our method on pre-trained LLMs such as LLaMA-3.2-1B, LLaMA-3.2-3B, LLaMA-3.2-3B-Instruct, LLaMA-3.1-8B, Qwen2.5-7B, Qwen2.5-Math-7B and Phi-4, evaluating on common generative, discriminative, math and reasoning tasks. 
We show that in most cases, CARVQ can achieve lower average bitwidth-per-parameter while maintaining reasonable perplexity and accuracy compared to scalar quantization.
%
Our contributions include a novel compression technique that is compatible with state-of-the-art transformer quantization methods
and can be seamlessly integrated into any hardware supporting 4-bit memory to reduce the model's memory footprint in memory-constrained devices. 
This work demonstrates a crucial step toward the efficient deployment of LLMs on edge devices.

\end{abstract}
\begin{figure}[ht]
\centering
\includegraphics[width=.95\linewidth]{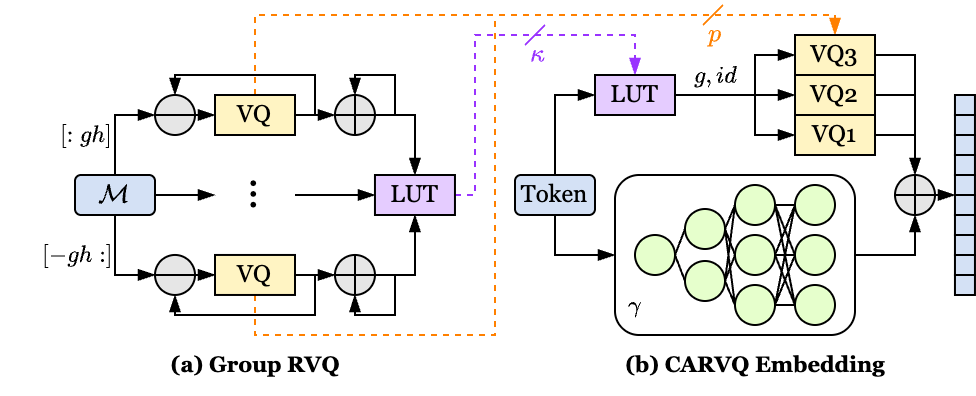}
\caption{\textbf{Overview of the CARVQ framework.} (a) The embedding matrix $\mathcal{M}$ is partitioned into 
groups, and each group is compressed using Residual Vector Quantization, with the results stored in a look-up table. LUT is stored in $\kappa$-bit and VQ elements are stored in original precision $p$. (b) At inference time, each token retrieves its corresponding quantized vectors from the look-up table. Simultaneously, it is processed by the corrective adaptor, a lightweight MLP. 
}
\label{fig:carvq}
\end{figure}
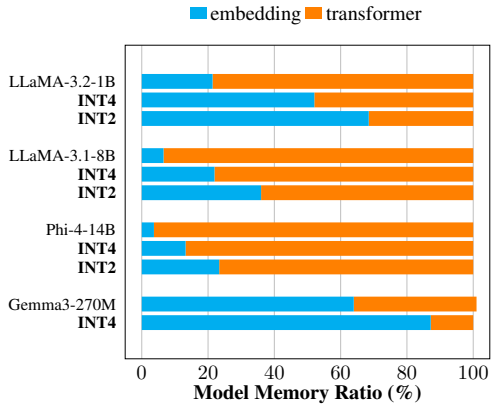
\begin{figure}[ht]
\centering
\begin{tikzpicture}[scale=0.7]
\begin{axis}[
    xbar stacked,
    enlargelimits=0.15,
    legend style={at={(0.5,1.15)},anchor=north,legend columns=2,draw=none},
    xlabel={\textbf{Model Memory Ratio (\%)}}, xmin=0, xmax=100,
    xmajorgrids=true,
    symbolic y coords={
    LLaMA-3.2-1B, INT4-1B, INT2-1B, 
    {}, LLaMA-3.1-8B, INT4-8B, INT2-8B, 
    {}, Phi-4-14B, INT4-Phi, INT2-Phi,
    {}, Gemma3-270M, INT4-Gemma
    },
    ytick=data,
    yticklabels={
    LLaMA-3.2-1B, \textbf{INT4}, \textbf{INT2},
    LLaMA-3.1-8B, \textbf{INT4}, \textbf{INT2},
    Phi-4-14B, \textbf{INT4}, \textbf{INT2},
    Gemma3-270M, \textbf{INT4}
    },
    y tick label style={rotate=0,anchor=east,font=\small,xshift=-1pt},
    ytick style={draw=none},
    xtick style={draw=none},
    y dir=reverse,
    xlabel style={at={(0.5,1ex)}},
    bar width = 8pt,
    y = 10pt,
    enlarge x limits=0.05,
    ]

\addplot+[xbar,draw opacity=0,fill=cyan] plot coordinates {
(21.36,LLaMA-3.2-1B) (52.06,INT4-1B) (68.48,INT2-1B)
(6.57,LLaMA-3.1-8B) (21.94,INT4-8B) (36,INT2-8B)
(3.67,Phi-4-14B) (13.22,INT4-Phi) (23.36,INT2-Phi)
(63.96,Gemma3-270M) (87.18,INT4-Gemma) 
};
\addplot+[xbar,draw opacity=0,fill=orange] plot coordinates {
(78.64,LLaMA-3.2-1B) (47.94,INT4-1B) (31.52,INT2-1B)
(93.43,LLaMA-3.1-8B) (78.06,INT4-8B) (64,INT2-8B)
(96.33,Phi-4-14B) (86.78,INT4-Phi) (76.64,INT2-Phi)
(37.04,Gemma3-270M) (12.82,INT4-Gemma) 
};
\legend{embedding, transformer}
\end{axis}
\end{tikzpicture}
\caption{With transformer layer quantization, the memory ratio of embedding layer increases compared to embedding layers left in FP16. 
}
\label{fig:model_ratio}
\end{figure}

\section{Introduction}

Transformer-based Large Language Models (LLM) are designed to handle extended contexts efficiently through processing tokens with attention mechanisms.
Transformer architecture can be primarily separated into three core components: embedding layer, transformer blocks, and prediction head. Each of these plays a distinct role in the overall processing pipeline, enabling the model to handle complex tasks across diverse domains.

While various compression techniques have been widely studied for transformer blocks, embedding layer compression has not been investigated extensively due to its simple nature of directly mapping from vocabulary tokens to high-dimensional vector representations. As most quantization are done to 4-bit datatypes, scalar quantization is often sufficient. We show in Figure~\ref{fig:model_ratio} that the portion of the embedding layer shrinks in larger models, but increases rapidly when post-training quantization (PTQ) operations are applied to the transformer layers. This ratio for INT4 quantized models are 52.06\% for LLaMA-3.2-1B~\cite{grattafiori2024llama}, 13.22\% for Phi-4 (14B)~\cite{abdin2024phi4technicalreport}, and 87.18\% for Gemma3-270M~\cite{gemmateam2025gemma3technicalreport}. Compressing the embedding layer thus has a more pronounced impact on smaller models, making them essential for deployment in resource-limited inference scenarios. Therefore, the embedding layer often becomes a bottleneck in resource-constrained environments as its relative memory footprint expands in compressed models, prompting further investigation into embedding layer compression.


In this work, we specifically focus on the deployment of LLMs on memory-constrained edge devices, where available memory is limited to a few GB. Such situations usually require the use of smaller (less than 8B), quantized models where the relative contribution of the full-precision (FP16) embedding layer to the total model memory is greater compared to larger models. Saving of a mere 0.5GB can grant 2B additional parameters at 4-bit precision or longer context-length, which would greatly improve the output performance. 

Quantization can be applied either during training in what is called quantization-aware training (QAT) or after during post-training quantization (PTQ).
While QAT remains the most effective for preserving the model's accuracy, it also presents some practical limitations. It requires access to the original training data, which is often private for LLMs. Secondly, QAT is computationally intensive as it requires retraining and fine-tuning of the model, often requiring a repeat of the specific training procedure of the model (e.g., instruction-tuning with RLHF). These constraints make QAT difficult to apply generally across diverse pre-trained models. On the other hand, PTQ offers a more generalizable and data-independent solution, as it can directly be applied to a frozen model. This is the solution adopted in this work.


To effectively target the embedding layer, we propose a novel compression method called CARVQ, a post-training Corrective Adaptor with Group RVQ. The proposed Corrective Adaptor ($\gamma$) relies on the composition of both linear and non-linear maps. More precisely, we define it as the composed map $\gamma=\sigma_1 \circ \sigma_0$, where $\sigma_0$ first embeds the tokens into a very small dimension $m$ and $\sigma_1$ expands the resulting vectors back to the embedding dimension $n$ through a multi-layer perceptron with small hidden dimensions. 
This Corrective Adaptor compensates the loss from group RVQ operation, which we use as an inexpensive strategy to retain the essential knowledge in the original embedding matrix. Through a careful decision of centroid bitwidth $\kappa$, CARVQ is orthogonal to existing transformer-layer quantization approaches such as activation-aware weight quantization (AWQ) without requiring any additional datatype support. 

To study the impact of CARVQ on pre-trained LLMs, we apply the proposed method on variations of three architectures, namely LLaMA-3.2-1B \cite{grattafiori2024llama}, LLaMA-3.2-3B \cite{grattafiori2024llama}, LLaMA-3.2-3B-Instruct \cite{grattafiori2024llama}, LLaMA-3.1-8B \cite{grattafiori2024llama}, Qwen2.5-7B \cite{qwen2.5}, Qwen2.5-Math-7B \cite{yang2024qwen25mathtechnicalreportmathematical}, and Phi-4 (14B) \cite{abdin2024phi4technicalreport}. We evaluate these models on four common types of NLP tasks: generative, discriminative, math, and reasoning.
In most cases, we observe that the model performance drop to be near-lossless at 2.4-bit bitwidth-per-parameter (bpp) and reasonable (perplexity  $<18$) 
at 1.6-bit bitwidth-per-parameter on average.

The main contributions are summarized below:
\begin{enumerate}
\item We introduce CARVQ, a novel post-training method for LLM embedding layer compression without requiring specialized hardware to support lower-bit storage. CARVQ combines group RVQ with Corrective Adaptor to maximize information retention in low bits.
\item We evaluate the proposed compression method on various task types, achieving better model perplexity and evaluation scores than the common approach of scalar quantization. CARVQ achieves 1.6 average-bitwidth-per-parameter compression on all models while scalar quantization does not hold model performance below 3 bits.
\item We demonstrate that CARVQ is compatible with transformer-layer quantization methods without requiring special datatype supports, unlike scalar quantization. 
This allows CARVQ to be readily fitted to most deployed LLMs today. We evaluate CARVQ on INT4-AWQ-quantized LLMs.

\end{enumerate}

\section{Related Works}





The rapid evolution of transformer-based architectures has significantly enhanced performance across natural language processing (NLP), computer vision (CV), and multimodal (MMMU) tasks. However, the massive computational and memory demands of these models remain a significant barrier to their deployment in resource-constrained environments such as edge devices and real-time applications. To address this issue, various compression techniques have been investigated.

\textbf{Architecture Preserving Compression} \cite{xiao2023smoothquant, lin2023awq, fang2023depgraph, ma2023llmpruner} aims to reduce the size of transformer models while preserving their overall structure and operational flow. These methods focus on minimizing redundant computations or parameters without altering the architecture itself. Quantization \cite{xiao2023smoothquant, lin2023awq,huijben2024residual,egiazarian2024extreme,van2024gptvq,dettmers2024qlora} converts high-precision weights and activations into lower-precision representations to reduce memory and computational costs. By carefully balancing precision loss and performance, quantization methods are particularly effective in deploying models on hardware with constrained resources, such as system-on-chip (SoC). Another complementary approach, pruning \cite{fang2023depgraph, ma2023llmpruner,ashkboos2024slicegpt}, eliminates weights or attention heads deemed less impactful based on predefined criteria. For example, head pruning and structured pruning have shown that many weights in transformer blocks contribute minimally to the overall accuracy. These methods allow significant reductions in size and computational demand while retaining the architectural integrity of the model.

\textbf{Architecture Adaptive Compression} \cite{hu2022lora, oseledets2011tensor} involves reconfiguring the model structure to achieve compression by replacing or simplifying specific layers. These methods embrace the notion that certain architectural modifications can provide significant efficiency gains while maintaining task accuracy. Low-Rank Adaptation (LoRA) \cite{hu2022lora} introduces low-rank parameter updates to large pre-trained weights, reducing memory footprint during fine-tuning on downstream tasks without the need to retrain. Another innovative method, Tensor Train Decomposition \cite{oseledets2011tensor}, decomposes large tensors into smaller low-rank tensors, significantly reducing memory footprint while maintaining accuracy.

\textbf{Embedding Layer Compression} \cite{xu2023tensorgpt, vincenti2024dynamic} specifically targets the reduction of parameters associated with embedding layers, which often constitute a substantial proportion of the overall model size, especially in multilingual architectures. In transformer models, embedding layers map input vocabulary and image-patch tokens to high-dimensional vectors, and their size grows proportionally with the vocabulary size. TensorGPT \cite{xu2023tensorgpt} leverages tensor train decomposition to represent these high-dimensional embeddings in a more compact manner, achieving parameter reduction with minimal degradation to model accuracy. However, the linear nature of TensorGPT results in accuracy drops in high-compression. Similarly, Dynamic Vocabulary Pruning \cite{vincenti2024dynamic} adjusts the vocabulary size adaptively based on the task or data requirements, pruning infrequently used tokens to reduce the embedding matrix dimensions. However, these methods are limited by the requirement of fine-tuning, lacking generalization.

\section{Background}
\subsection{Input embedding matrix $\mathcal{M}$}
	We denote by $\mathcal{W}$ the set of all tokens, known as the vocabulary, for a given alphabet.
	
	\begin{definition} An embedding is a mapping from $\mathcal{W}$ to $\mathbb{R}^n$ for $n \geq 1$. It takes as input a token $T$ and returns an $n$-dimensional real vector, that is $\sigma : \mathcal{W} \rightarrow \mathbb{R}^n$ sends $T$ to $(v_1, v_2, \cdots, v_n)$, where $v_i \in \mathbb{R}$.
	\end{definition}
The embedding map is learned during training where weights of the following matrix are given.

\begin{definition}
Let $V$ denote the number of tokens in a given vocabulary, \emph{i.e.,} we have $V = |\mathcal{W}|$. The embedding weight matrix is the matrix $\mathcal{M} \in \mathbb{R}^{V \times n}$ where each row $\mathcal{M}_i$, for $i = 1, \cdots, V$ of the matrix corresponds to $\mathcal{M}_i = \sigma(T_i)$ for $T_i \in \mathcal{W}$. 
\end{definition}

The number of coefficients $V \times n$ in the embedding weight matrix corresponds to the number of parameters in the model. 
Initially, the embedding coefficients in the matrix $\mathcal{M}$ are set to random real values. 
During the training process, these coefficients are updated through backpropagation as the model analyzes more and more data. 
Token embeddings can be pre-trained using traditional algorithms such as skip gram~\cite{guthrie2006closer} or CBOW~\cite{mikolov2013efficient}. However, for task-specific foundational LLM model fine-tuning, it is usually preferable to adapt the token embedding to the data distribution considered and the semantics of the specific downstream tasks.
\subsection{Group Residual Vector Quantization}\label{sec:RVQ}


Residual Vector Quantization~\cite{chen2010approximate} (RVQ) can be used to compress the embedding layer in an LLM by representing the $n$-dimensional embedding vectors using a sequence of lower-dimensional quantized residuals. More specifically, RVQ works by iteratively applying vector quantization~\cite{gray1984vector} (VQ) and encoding the residuals.

Vector quantization represents a high-dimensional vector with the closest centroid from a pre-trained codebook $\mathcal{C} \in \mathbb{R}^{K \times n}$ defined as a set of $K$ centroids, \emph{i.e.,} $n$-dimensional vectors.
First, the codebook is trained using for example $K$-mean clustering algorithm on the entire embedding matrix. 
For each embedding vector $\sigma(T_i) \in \mathbb{R}^n$, one can compute the distance $\hat{j} = \text{arg}\min_j || \sigma(T_i) - c_i ||_2$
from it to each centroid $c_i \in \mathcal{C}$ and identify the closest one. 
The embedding vector $\sigma(T_i)$ is now represented by the index $\hat{j}$ of the closest centroid $c_i \in \mathcal{C}$. During inference, the embedding vector $\sigma(T_i)$ is reconstructed as an approximation of the original vector with some quantization error, called residual, defined by the quantity $r_{i} = \sigma(T_i) - c_{i}$, computed for all $T_i \in V$. 
This process can be repeated iteratively, say $L$ times, and one can then apply vector quantization on the residuals obtained from the previous step using a new codebook $\mathcal{C'}$, resulting in a new set of residuals $r'_{i} = r_{i} - c'_i$.
If one repeats this process where at each iteration a different codebook is used, each embedding vector $\sigma(T_i)$ is now represented by a vector of length $L$ which represents the sequence of $L$ quantized indices $(j_{i,1}, \cdots, j_{i,L})$ where $j_{i, j}$ is the index of the closest centroid in the codebook $\mathcal{C}_j$ at layer $j$ for the embedding vector $\sigma(T_i)$. 
In order to reconstruct the embedding vector $\sigma(T_i)$ one can sum the centroids selected by the indices $(j_{i, 1}, \cdots, j_{i, L})$. The goal of RVQ is to minimize the reconstruction error while still keeping the storage cost as small as possible. 

\paragraph{Group RVQ.} Group RVQ was introduced in~\cite{yang2023hifi} and offers a variant of RVQ where the set of embedding vectors is divided into groups and RVQ is applied to each group separately as described above. The
group RVQ outputs are then combined to obtain the final quantization results. 
In this work, we will consider this group RVQ method as it is expected to have smaller reconstruction error than standard RVQ since it operates over smaller sets.

\paragraph{Compression with group RVQ.}
Residual Vector Quantization is used to reduce the storage size of the embedding matrix but does not effectively reduce the number of parameters in the embedding layer. Let us analyze how the storage size is compressed when using group RVQ with group size $g$ with sub-vector dimension $h$. We initially have $n$-dimensional embedding vectors for each vocabulary in $V$ resulting in an embedding matrix of size $n \times V$. We start by splitting the input set of embedding vectors into $nV/gh$ groups such that each group contains $g$ embedding vectors of dimension $h$ for any $g, h > 0$ such as $gh|nV$. We then consider the bit-compression ratio for \textit{a single group}.
We will compute the following ratio:
\[
\frac{\text{storage emb. matrix}}{\text{storage comp. model}} = \frac{gh\times p}{\text{storage comp. model}}
\]
where $p$ is the precision of the original embedding coefficients.
%
Let us now count the storage size in the compression model. There are two elements that need to be stored. The quantized indices and the codebooks.
The total number of centroids is $2^{\kappa}$, where $\kappa$ is the number of bits necessary to index a centroid, and each centroid is of length~$h$. The resulting codebook storage for $L$ iterations is then equal to $Lh2^{\kappa} \times p$ bits.
For the quantization indices, each of the $g$ embedding vectors goes through $L$ iterations. For each iteration, an index is stored into a $2^{\kappa}$-sized codebook, requiring $\kappa$ bits.
Finally, we have the compression ratio
\[
\text{compression-ratio}_{\text{bits}} = \frac{gh \times p}{Lh2^\kappa \times p + gL\kappa}.
\]
In our work, we will focus on the average bitwidth-per-parameter representing how many bits per original parameter we are now effectively using after compression. This quantity is computed as
$$B_{\text{RVQ}}= \frac{p}{\text{comp-ratio}_{\text{bits}}}=p\times\frac{Lh2^\kappa \times p + gL\kappa}{gh\times p}.$$

\usetikzlibrary{shapes.geometric}

\begin{figure}[t
]
    \centering
    \pgfplotsset{
        tick label style={font=\small},
        label style={font=\small},
        legend style={font=\small}, 
        xlabel style={at={(0.5,1ex)}},
        ylabel style={at={(3ex,0.5)}}
    }
    \pgfdeclareplotmark{redstar}{
        \node[star,star point ratio=2.25,minimum size=8pt,
              inner sep=0pt,draw=orange,solid,fill=orange] {};
    }
    \pgfdeclareplotmark{bluestar}{
        \node[star,star point ratio=2.25,minimum size=8pt,
              inner sep=0pt,draw=cyan,solid,fill=cyan] {};
    }
    \pgfdeclareplotmark{redsquare}{
        \node[rectangle,minimum size=6pt,
              inner sep=0pt,draw=red,solid,fill=red] {};
    }
    \pgfdeclareplotmark{bluesquare}{
        \node[rectangle,minimum size=6pt,
              inner sep=0pt,draw=blue,solid,fill=blue] {};
    }
    \begin{tikzpicture}[scale=0.7]
        \begin{axis}[
          ylabel=Perplexity,
          ymajorgrids=true,
          ytick distance=2,
          xlabel=Average Bitwidth,
          xmin=1, xmax=5,
          ymin=6, ymax=18,
          legend plot pos=right,
          ytick style={draw=none},
          xtick style={draw=none},
        ]
          \addplot[mark=bluesquare,blue,line width=0.5mm] coordinates {
            (2,154)
            (3,8.58)
            (4,7.92)
            (16, 7.81)
          };
          \addlegendentry{LLaMA-3.2-3B SQ}
          \addplot[mark=bluestar,cyan,line width=0.5mm] coordinates {
            (1.655,16.34)
            (2.405,8.43)
            (3.155,7.97)
            (16, 7.81)
          };
          \addlegendentry{LLaMA-3.2-3B CARVQ}
          \addplot[mark=redsquare,red,line width=0.5mm,dashed] coordinates {
            (2,8.65)
            (3,6.33)
            (4,6.26)
            (16, 6.24)
          };
          \addlegendentry{LLaMA-3.1-8B SQ}
          \addplot[mark=redstar,orange,line width=0.5mm,dashed] coordinates {
            (1.633, 7.41)
            (2.383, 6.55)
            (3.133, 6.43)
            (16, 6.24)
          };
          \addlegendentry{LLaMA-3.1-8B CARVQ}
        \end{axis}
    \end{tikzpicture}
\caption{Wikitext-2 Perplexity for LLaMA-3.2-3B and LlaMA-3.1-8B with scalar quantization and CARVQ.}
\label{fig:comparison}
\end{figure}
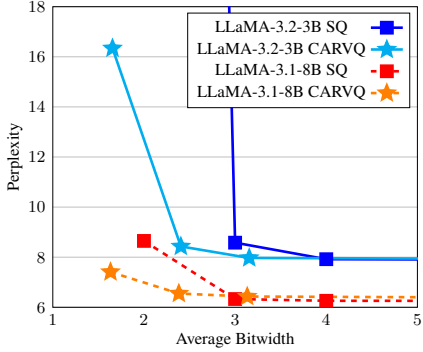

\section{CARVQ Embedding Compression}
\label{section:embedding_layer_compression}

In this section, we introduce a novel embedding CARVQ, defined as the composition of two methods: Corrective Adaptor (\ref{sec:composed_emb}) and Group RVQ (\ref{sec:carvq_grvq}). Corrective Adaptor works as non-linear intermediate maps $\sigma_1$ expanding the small dimension output of linear map $\sigma_0$ to the original embedding dimension. Through this design we can efficiently compensate the precision loss from Group RVQ operation. Figure~\ref{fig:carvq} illustrates CARVQ's framework.

\subsection{Corrective Adaptor}\label{sec:composed_emb}
Corrective Adaptor (CA) performs contraction-expansion strategy to significantly reduce the number of parameters required to map a token to its embedding.
First, let us define the following map to map a token $\mathcal{W}$ to a small-dimension vector
$\sigma_0 \colon \mathcal{W} \to \mathbb{R}^m$ sending
    	$T \mapsto (v_1, v_2, \cdots, v_m)$
where $m \ll n$. We will call index the output of $\sigma_0$. We will refer to the dimension $m$ as the \textit{corrective width}. Then, we define  $\sigma_1 \colon \mathbb{R}^m \to \mathbb{R}^n$ sending
       $ (v_1, v_2, \cdots, v_m) \mapsto (\tilde{v}_1, \tilde{v}_2, \cdots, \tilde{v}_n).$
This map expands the $m$-dimensional vectors to the embedding dimension $n$.
Note that the number of parameters in the model is then equal to $mV + nm$, and as $V$ and $n$ are usually fixed by the dataset and model considered, the corrective width $m$ is the hyper-parameter introduced by our new method that will change the average bitwidth-per-parameter. 

\paragraph{Defining the map $\sigma_0$.} Initially, for a given corrective width $m$, the coefficients of the vectors $\sigma_0(T) = (v_1, \cdots, v_m)$ are randomly assigned for each token $T$ in the vocabulary. These coefficients are then updated via training against the original embedding matrix $\mathcal{M}$.

\paragraph{Defining the map $\sigma_1$.} The question now remains as to how one defines the map $\sigma_1$.
We define $\sigma_1$ as a multi-layer perceptron, a composition of linear and non-linear functions $h_L$ and $h_{NL_i}$, respectively. This means we have the following composed embedding $\sigma_1 = h_L \circ h_{NL_k} \circ \cdots \circ h_{NL_1}$, where the non-linear maps $h_{NL_i}$ for $i = 1, \cdots, k$, are defined as 
$h_{NL_i} \colon \mathbb{R}^{m_i} \to \mathbb{R}^{m_{i+1}}$ sending 
	$x \mapsto \text{Relu}(W_i \cdot x + b_i)$
with $W_i \in \mathbb{R}^{m_{i+1} \times m_i}$ and $b_i \in \mathbb{R}^{m_{i+1}}$. Note that $m_1 = m$.
The last function $h_L$ is linear and corresponds to a weighted summation
	$h_L \colon \mathbb{R}^{m_{k+1}} \to \mathbb{R}^{n}$ sending
	$x \mapsto W_L \cdot x + b_L$
with $W_L \in \mathbb{R}^{n \times m_{k+1}}$ and $b_L \in \mathbb{R}^{n}$. The last embedding $h_L$ must be linear so that the token embedding can have both negative and positive values in its vector representation. 
We add layer normalization between dense layers with ReLU activation to facilitate training.
The dimensions $m_i$ for $i = 1, \cdots, k+1$ are experimentally chosen and fine-tuned taking into account downstream tasks. Moreover, the number $k$ of non-linear maps to apply can also vary.
The number of model parameters in this case is equal to \[N_P=mV + \sum_{i=1}^{k} m_i m_{i+1} + \sum_{i=1}^k m_{i+1} + m_{k+1}n + n \]
where the values $m_im_{i+1}$ are the number of parameters in the matrices $W_i$, the $m_{i+1}$ are the number of parameters in the $b_i$, $mV$ is the size of $\mathcal{M}_{\sigma_0}$ as before, $nm_{k+1}$ and $n$ are the sizes of $W_L$ and $b_L$. 


\paragraph{Compression ratio.} By introducing an intermediate mapping that operates on much smaller vectors of dimension $m$, we were able to compress our model by the following ratio
\begin{equation*}
\resizebox{\hsize}{!}{
$\frac{nV}{mV + \sum_{i=1}^{k} m_i m_{i+1} + \sum_{i=1}^k m_{i+1} + m_{k+1}n + n }$
}
\end{equation*}
Parameters corresponding to biases are nearly negligible compared to weight parameter counts.

\begin{table*}[ht]
\centering
\resizebox{\linewidth}{!}{
\begin{tabular}{cc*{5}{cc}|cc}
\hline
\multirow{2}{*}{\textbf{Prec.}} &
\multirow{2}{*}{\textbf{Method}} & 
\multicolumn{2}{c}{\textbf{LLaMA-3.2-1B}} &
\multicolumn{2}{c}{\textbf{LLaMA-3.2-3B}} &
\multicolumn{2}{c}{\textbf{LLaMA-3.2-3B-Inst}} &
\multicolumn{2}{c}{\textbf{LLaMA-3.1-8B}} &
\multicolumn{2}{c}{\textbf{Qwen2.5-7B}} &
\multicolumn{2}{|c}{\textbf{Mean}} \\
&& Prec. & PPL &
Prec. & PPL &
Prec. & PPL &
Prec. & PPL &
Prec. & PPL &
Prec. & $\Delta$ PPL \\
\hline
\hline

16 & FP16 &
16 & 9.75 &
16 & 7.81 &
16 & 11.05 &
16 & 6.24 &
16 & 6.85 &
16 & 0 \\
\hline

$4\pm0.5$ & INT4 &
4 & 9.98 &
4 & 7.92 &
4 & 11.18 &
4 & 6.26 &
4 & 6.85 &
4 & 0.098 \\
\hline

\multirow{2}{*}{$3\pm0.5$} & INT3 &
3 & 11.42 &
3 & 8.58 &
3 & 12.26 &
3 & \textbf{6.33} &
3 & \textbf{6.86} &
3 & 0.750 \\
& CARVQ-4 &
3.201 & \textbf{10.15} &
3.155 & \textbf{7.97} &
3.155 & \textbf{11.48} &
3.133 & 6.43 &
3.131 & \textbf{6.86} &
3.155 & \textbf{0.238} \\
\hline

\multirow{3}{*}{$2\pm0.5$} & INT2 &
2 & 181 &
2 & 154 &
2 & 110 &
2 & 8.65 &
2 & 7.44 &
2 & 83.88 \\
& CARVQ-3 &
2.451 & \textbf{10.81} &
2.405 & \textbf{8.43} &
2.405 & \textbf{11.70} &
2.383 & \textbf{6.55} &
2.381 & \textbf{6.87} &
2.405 & \textbf{0.532} \\
& CARVQ-2 &
1.701 & 14.27 &
1.655 & 16.34 &
1.655 & 14.49 &
1.633 & 7.41 &
1.631 & 6.91 &
1.655 & 3.544 \\
\hline
\end{tabular}}
\caption{CARVQ Results on Text Generation for Wikitext-2. LLaMA-3.2-1B (128256x2048), LLaMA-3.2-3B (128256x3072), LLaMA-3.1-8B (128256x4096), Qwen2.5-7B (152064x3584). CARVQ-$L$ represents $L$-iteration RVQ. For CARVQ, the centroid count $K$ is fixed to 16 (4-bit), the sub-vector dimension $h$ to 8, and the group size $g$ to 1024. Precision in the table above corresponds to average bitwidth-per-parameter $B$. 
}
\label{table:delta_results2}
\end{table*}

\begin{table}[t]
\centering
\resizebox{1\columnwidth}{!}{
\begin{tabular}{cccc}
\hline
Method & $B_{RVQ}$ & $B_{CA}$ & mem. gain (GB) \\
\hline
\hline
CARVQ-4 & 3.000 & 0.155 & 0.756 \\
CARVQ-3 & 2.250 & 0.155 & 0.831 \\
CARVQ-2 & 1.500 & 0.155 & 0.877 \\
\hline
\end{tabular}}
\caption{Memory overhead of the corrective adaptor is sufficiently small compared to the group RVQ for the overall compression ratio to be dictated by the group RVQ compression.}
\label{tab:memory-saving}
\end{table}

\begin{table}[t]
\centering
\resizebox{1\columnwidth}{!}{
\begin{tabular}{cccc}
\hline
Model & Model (GFLOPS) & CA (MFLOPS) & CA (MB) \\
\hline
\hline
LLaMA-3.2-1B & 0.62 & 0.63 & 2.52 \\
LLaMA-3.2-3B & 1.61 & 0.89 & 3.56 \\
LLaMA-3.1-8B & 3.75 & 1.15 & 4.62 \\
Qwen-2.5-7B & 3.53 & 1.02 & 4.08 \\
Phi-4 (14B) & 7.08 & 1.41 & 5.66 \\
\hline
\end{tabular}}
\caption{Corrective adaptor computational overhead is at most 0.1\% of the original model with configurations described in Section~\ref{sec:experiments:implementation}.}
\label{tab:overhead}
\end{table}

\subsection{Combining Corrective Adaptor with group RVQ}
\label{sec:carvq_grvq}

For the final compression of the embedding matrix, we combine our Corrective Adaptor method with group RVQ explained in Section~\ref{sec:RVQ}. 
Group RVQ retains knowledge in the original embedding matrix of the pre-trained model with a minimal number of bits without requiring any special datatype support in hardware. As the original datasets for most models are not available during compression, this PTQ approach allows retention of knowledge, especially when leveraging Corrective Adaptor to minimize the precision loss between the output of original embedding $\mathcal{M}$ and CARVQ. Without the Corrective Adaptor, group RVQ would be too destructive, significantly impacting the performance of the model. Their combination allows an inexpensive method minimizing the reconstruction loss without any fine-tuning. 

We now concretely explain how we combine these methods.
We start by reshaping the embedding matrix. We divide each embedding vector (\emph{i.e.,} each row of $\mathcal{M}$) into $j$ sub-vectors each of dimension $h$, where naturally $jh= n$.
The embedding matrix $\mathcal{M}$ is then reshaped into matrix $\mathcal{M}'$ of size $nV/h \times h$ where each row now corresponds to an $h$-dimensional sub-vector. The reason we reshape the matrix is to consider RVQ on smaller vectors of dimension $h \ll n$ for which similarity search (e.g., nearest neighbor search) is more effective. In addition to this reshaping, as we consider group RVQ, we split $\mathcal{M}'$ into $nV/gh$ groups such that each group corresponds to a matrix of size $g \times h$ which will be compressed using RVQ.

Recall that $L$ is the number of iterations considered in RVQ, $2^\kappa$ is the number of centroids, and for each group of size $g$, each sub-vector of dimension $h$, after applying RVQ, is represented as an $L$-dimensional vector $(j_1, j_2, \cdots, j_L)$ of indices.
From Section~\ref{sec:RVQ}, we already have the average bitwidth-per-parameter after applying group RVQ referred to as
$B_{\text{RVQ}}$. 
Let us now analyze the average bitwidth-per-parameter when combining group RVQ with our Corrective Adaptor. We have already described in Section~\ref{sec:composed_emb} the parameter count $N_p$ corresponding to our embeddings $\sigma_0$ and $\sigma_1$. Therefore, we define
$B_{\text{CA}}=p\times\frac{N_p}{nV},$
to be the average bitwidth-per-parameter resulting from our Corrective Adaptor compression.
The average bitwidth-per-parameter considering group RVQ and Corrective Adaptor is equal to 
$B_{\text{CARVQ}}=B_{\text{CA}}+B_{\text{RVQ}}.$
In most scenarios, the number of parameters in RVQ is much greater than the corrective layer, meaning $B_{CA} \ll B_{RVQ}$. We show this in Table~\ref{tab:memory-saving}. This shows that, as long as the corrective width $m$ is kept relatively low, the hidden dimension of the intermediate map $\sigma_1$ is not too relevant. We also find the computational overhead to be at most 0.1\% of the entire model, as shown in Table~\ref{tab:overhead}.

\begin{table*}[ht]
\centering
\resizebox{\linewidth}{!}{
\begin{tabular}{cc|*{1}{cc}|*{1}{cc}|*{2}{cc}|*{3}{cc}}
\hline
\multirow{2}{*}{\textbf{Model}} & 
\multirow{2}{*}{\textbf{Dataset}} &
\multicolumn{2}{c}{\textbf{FP16}} &
\multicolumn{2}{|c}{\textbf{INT4}} &
\multicolumn{2}{|c}{\textbf{INT3}} &
\multicolumn{2}{c}{\textbf{CARVQ-4}} &
\multicolumn{2}{|c}{\textbf{INT2}} &
\multicolumn{2}{c}{\textbf{CARVQ-3}} &
\multicolumn{2}{c}{\textbf{CARVQ-2}} \\
&& Prec. & Acc. &
Prec. & Acc. & 
Prec. & Acc. & 
Prec. & Acc. & 
Prec. & Acc. & 
Prec. & Acc. & 
Prec. & Acc. \\
\hline
\hline

\multirow{3}{*}{LLaMA-3.2-1B} 
& Hella & 
\multirow{3}{*}{16} & 47.72 & 
\multirow{3}{*}{4} & 47.66 & 
\multirow{3}{*}{3} & 46.83 & 
\multirow{3}{*}{3.201}& \textbf{47.09} & 
\multirow{3}{*}{2} & 30.43 & 
\multirow{3}{*}{2.451} & \textbf{45.86} &
\multirow{3}{*}{1.701} & 40.58 \\
& Wino & 
& 60.69 & 
& 60.85 & 
& \textbf{60.46} & 
& 59.98 & 
& 54.93 & 
& \textbf{60.30} &
& 56.04 \\
& Piqa & 
& 74.48 & 
& 74.70 & 
& 73.23 & 
& \textbf{73.76} & 
& 60.83 & 
& \textbf{72.31} &
& 70.46 \\
\hline

\multirow{3}{*}{LLaMA-3.2-3B} 
& Hella & 
\multirow{3}{*}{16} & 55.31 & 
\multirow{3}{*}{4} & 54.22 & 
\multirow{3}{*}{3} & 50.46 & 
\multirow{3}{*}{3.155}& \textbf{55.38} & 
\multirow{3}{*}{2} & 35.15 & 
\multirow{3}{*}{2.405} & \textbf{54.63} &
\multirow{3}{*}{1.655} & 50.33 \\
& Wino & 
& 69.85 & 
& 68.98 & 
& \textbf{69.38} & 
& 68.67 & 
& 56.83 & 
& \textbf{67.80} &
& 64.48 \\
& Piqa & 
& 76.66 & 
& 76.39 & 
& 75.57 & 
& \textbf{76.88} & 
& 67.57 & 
& \textbf{76.33} &
& 74.59 \\
\hline

\multirow{3}{*}{LLaMA-3.1-8B} 
& Hella & 
\multirow{3}{*}{16} & 60.01 & 
\multirow{3}{*}{4} & 60.07 & 
\multirow{3}{*}{3} & \textbf{60.00} & 
\multirow{3}{*}{3.133}& 59.83 & 
\multirow{3}{*}{2} & 54.74 & 
\multirow{3}{*}{2.383} & \textbf{59.41} &
\multirow{3}{*}{1.633} & 55.11 \\
& Wino & 
& 73.88 & 
& 73.72 & 
& 73.24 & 
& \textbf{73.72} & 
& 70.00 & 
& \textbf{73.24} &
& 70.32 \\
& Piqa & 
& 80.14 & 
& 79.82 & 
& 79.33 & 
& \textbf{79.82} & 
& 75.57 & 
& \textbf{79.33} &
& 77.15 \\
\hline

\multirow{3}{*}{Qwen2.5-7B} 
& Hella & 
\multirow{3}{*}{16} & 60.02 & 
\multirow{3}{*}{4} & 59.97 & 
\multirow{3}{*}{3} & \textbf{60.03} & 
\multirow{3}{*}{3.131}& 59.90 & 
\multirow{3}{*}{2} & 59.11 & 
\multirow{3}{*}{2.381} & 59.98 &
\multirow{3}{*}{1.631} & \textbf{60.05} \\
& Wino & 
& 72.93 & 
& 73.32 & 
& \textbf{72.69} & 
& \textbf{72.69} & 
& 70.80 & 
& \textbf{72.14} &
& 71.11 \\
& Piqa & 
& 78.73 & 
& 78.45 & 
& \textbf{78.73} & 
& 78.62 & 
& \textbf{78.89} & 
& 78.56 &
& 78.78 \\
\hline 

\multirow{1}{*}{Qwen2.5-Math-7B} 
& GSM8K &
\multirow{1}{*}{16} & 83.62 & 
\multirow{1}{*}{4} & 83.78 & 
\multirow{1}{*}{3} & 83.02 & 
\multirow{1}{*}{3.131}& \textbf{84.08} & 
\multirow{1}{*}{2} & 47.54 & 
\multirow{1}{*}{2.381} & \textbf{83.70} &
\multirow{1}{*}{1.631} & 83.24 \\
\hline

\multirow{3}{*}{Phi-4 (14B)} 
& GSM8K &
\multirow{3}{*}{16} & 88.78 & 
\multirow{3}{*}{4} & 88.25 & 
\multirow{3}{*}{3} & 88.02 & 
\multirow{3}{*}{3.138}& \textbf{88.63} & 
\multirow{3}{*}{2} & \textbf{88.55} & 
\multirow{3}{*}{2.388} & 88.10 &
\multirow{3}{*}{1.638} & 87.79 \\
& ARC\textsubscript{C} & 
& 55.55 & 
& 55.72 & 
& \textbf{56.74} & 
& 54.86 & 
& 55.03 & 
& 55.20 &
& \textbf{55.38} \\
& ARC\textsubscript{E} & 
& 81.48 & 
& 81.61 & 
& \textbf{81.90} & 
& 81.65 & 
& 82.24 & 
& 81.82 &
& 80.43 \\
\hline
\hline

\multicolumn{2}{c|}{\textbf{Avg. Precision} / \textbf{Mean $\Delta$ Acc.}}
& 16 & - & 
4 & -0.15 &
3 & -0.64 &
3.148 & \textbf{-0.27} &
2 & -8.23 &
2.398 & \textbf{-0.70} &
1.648 & \underline{-2.75} \\
\hline

\end{tabular}}
\caption{Comparison of accuracy on discriminative and math/reasoning tasks. Phi-4's embedding matrix is of size (100352$\times$5120). Exact-match accuracy with flexible extraction is reported for GSM8K. ARC\textsubscript{C} and ARC\textsubscript{E} represent ARC challenge and easy tasks, respectively. CARVQ-$L$ represents $L$-iteration RVQ. For CARVQ, the centroid count $K$ is fixed to 16 (4-bit), the sub-vector dimension $h$ to 8, and the group size $g$ to 1024. Precision in the table above corresponds to average bitwidth-per-parameter $B$. }
\label{table:delta_eval}
\end{table*}

\section{Experimental results}\label{sec:experiments}

\subsection{Implementation Details}
\label{sec:experiments:implementation}

\paragraph{LLM Evaluation.} 
We evaluate our method on different architectures: LLaMA-3.2-1B \cite{grattafiori2024llama}, LLaMA-3.2-3B \cite{grattafiori2024llama}, LLaMA-3.2-3B-Instruct \cite{grattafiori2024llama}, LLaMA-3.1-8B \cite{grattafiori2024llama}, Qwen2.5-7B \cite{qwen2.5}, Qwen2.5-Math-7B \cite{yang2024qwen25mathtechnicalreportmathematical}, and Phi-4 (14B) \cite{abdin2024phi4technicalreport}. All pretrained models are open-source from HuggingFace, with our Corrective adaptor implemented in PyTorch. We do not consider weight-tying \cite{press2017using} in our work due to its limited usage in recent high-complexity tasks. We assess four types of LLM tasks for evaluation: generative, discriminative, math, and reasoning. For the generative task, we evaluate perplexity on Wikitext-2 \cite{merity2016pointer} following previous quantization works \cite{lin2023awq, chen2025hpca}. Discriminative tasks are evaluated with three benchmarks: HellaSwag \cite{zellers2019hellaswag}, WinoGrande \cite{ai2:winogrande}, and Piqa \cite{Bisk2020}. Math tasks are evaluated with GSM8K \cite{cobbe2021gsm8k}. We evaluate on ARC Challenge and ARC Easy \cite{Clark2018ThinkYH} for the reasoning task. 

\paragraph{Corrective Adaptor Implementation.}
Our method consists of two components: group RVQ and our Corrective Adaptor method. We acquire RVQ by applying K-Means clustering at each iteration for each group with a tolerance of $1e^{-4}$. 
Corrective Adaptor hyper-parameters are $[m_1,m_2,m_3]=[16,384,512]$ for all models.
Note that we promote overfitting during training as we want the network output to be as close as possible to the original embedding vector.

\paragraph{Training.}

As the input to embedding layers are a scalar pointing to a row in the embedding matrix, the data used to train the Corrective Adaptor is limited to the input dimension of embedding matrix. This also means that we can overfit the MLP as the input is fixed to a specific set, further lowering the expressiveness requirement of the adaptor. Then, with roughly 150,000 data points (English), we train such that the L1 loss of the final output sum of RVQ and Corrective Adaptor with the original embedding vector for each vocabulary is minimized. We train for 500 iterations with Adam for learning rate of 1e-3 on a RTX 4090 for 2 minutes.

\paragraph{Quantization Scheme for Comparison.}
To our knowledge, there are no open-source post-training compression methods targeting embedding layers. Thus, we compare Corrective Adaptor with scalar quantization, as prior quantization works \cite{lin2023awq, liu2024spinquant} utilize scalar quantization for the embedding layer. To ensure fair comparison, we keep the transformer layers in their original precision (FP16/BF16). 
Although scalar quantization can achieve a slightly lower average error per dimension, its uniform bins make it highly sensitive to large‐magnitude embedding values, producing a skewed error distribution. In contrast, group RVQ’s iterative residual steps produce errors that are approximately zero‐mean and more concentrated. Such zero‐centered, low‐variance errors are better suited for being corrected by the corrective adaptor. 

\subsection{Evaluation Details}
\paragraph{Generative Tasks.}
Table \ref{table:delta_results2} presents perplexity changes when applying different post-training compression methods to the embedding matrix, namely scalar quantization and our CARVQ-$L$ method for $L$-iteration RVQ. For 4, 3 and 2-layer CARVQ, we observe a $0.238$, $0.532$, and $3.544$ perplexity increase on average compared to the FP16 baseline models. On the other hand, scalar quantization works well up to 3-bit quantization with average perplexity increase of $0.098$ and $0.75$. However, 2-bit quantization fails to capture enough information for most models with an average perplexity loss of $84$. In comparison, the proposed CARVQ is very stable at 3 iterations with an average bit-per-parameter of 2.405, even preserving most of the structure at 2 iterations with 1.655 bit-per-parameter. We sometimes also notice marginal improvements ($<1\%$) over baseline when CARVQ is applied. We visualize the comparison in Fig. \ref{fig:comparison} with LLaMA-3.2-3B and LLaMA-3.1-8B. Furthermore, compared to INT3 and INT2 quantization, which require hardware optimization due to limited architecture support, CARVQ-3 and CARVQ-2 only utilize 4-bit and 16-bit datatypes for storage, making them compatible with all recent architectures supporting INT4. By adapting the centroid bitwidth $\kappa$, we can adapt CARVQ to any appropriate hardware. This demonstrates that the combination of group RVQ and corrective adaptors in CARVQ significantly reduces quantization error, achieving low-bit embedding without sacrificing hardware compatibility.

\paragraph{Discriminative Tasks.}
Table \ref{table:delta_eval} exhibits a comparison of evaluation accuracy on discriminative tasks with CARVQ and scalar quantization applied. With $L$=2, 3 and 4, we attain mean accuracy loss of 0.34, 0.88, and 3.45 for discriminative tasks, showing decent performances even at average bitwidth-per-parameter of 1.68. In comparison, scalar quantization mean losses are at 0.87 for INT3 and 7.97 for INT2, so 2-bit quantization would risk significant loss in answer quality. Moreover, this bigger gap is more pronounced in smaller models such as LLaMA-3.2-1B, where the accuracy gap between INT2 and CARVQ-2 is more than $10\%$. This shows that CARVQ can achieve much smaller bidwidth-per-parameter while maintaining good accuracy compared to scalar quantization for this task. 

\paragraph{Math Tasks.}
Table \ref{table:delta_eval} also compares the accuracy on math tasks with Qwen2.5-Math-7B and Phi-4. For both models, CARVQ methods result in an average minimal accuracy loss of less than $0.5\%$. However, we see a sudden loss of accuracy in INT3 and INT2, each of $0.6\%$ and $36\%$. As mathematical prompts require high retention of reasoning and memory, scalar quantization to INT2 likely simplifies the complex math problems too much. However, we observe CARVQ holding accuracy even at 1.63 bits, likely due to improved precision withheld from storing the VQ vectors in the original precision of FP16.

\paragraph{Reasoning Task.}
We include reasoning tasks evaluation with Phi-4 to demonstrate further generalization to different tasks in Table \ref{table:delta_eval}. As Phi-4 is larger at 14B parameters, we see retention of performance even with low-bit quantization, even showing improvements in both easy and challenge evaluations. Overall, both scalar quantization and CARVQ quantization result in a minimal ($<1.1\%$) drop in accuracy under reasoning tasks when a sufficiently large model is used to counter the quantization loss in the embedding layer.
Through extensive evaluation across various tasks, we conclude CARVQ-3 offers the best trade-off between accuracy retention and compression ratio.

\begin{table}[t!]
\centering

\resizebox{1\columnwidth}{!}{
\begin{tabular}{ccccc}
\hline
\multirow{2}{*}{\textbf{Method}} & 
\multicolumn{2}{c}{\textbf{LLaMA-3.2-3B}} & 
\multicolumn{2}{c}{\textbf{Qwen2.5-7B}} \\
& Prec. & PPL & Prec. & PPL \\
\hline
\hline
FP16 & 16 & 7.81 & 16 & 6.85 \\
\hline
CA+INT3 & 3.155 & \textbf{7.90} & 3.131 & \textbf{6.86} \\
CARVQ-4 & 3.155 & 7.97 & 3.131 & \textbf{6.86} \\
\hline
CA+INT2 & 2.155 & 12.51 & 2.131 & 7.39 \\
CARVQ-3 & 2.405 & \textbf{8.43} & 2.381 & \textbf{6.87} \\
\hline
CA+INT1 & 1.155 & 14528 & 1.131 & 46480 \\
CARVQ-2 & 1.655 & \textbf{16.34} & 1.631 & \textbf{6.91} \\
CARVQ-2 (k=3) & 1.155 & 3230 & - & - \\
\hline
\end{tabular}}
\caption{Ablation of quantization with scalar quantization. We evaluate the perplexities of LLaMA-3.2-3B and Qwen2.5-7B on Wikitext-2.} 
\label{tab:quant_ablation}
\end{table}

\begin{table}[t]
\centering

\resizebox{1\columnwidth}{!}{
\begin{tabular}{ccc}
\hline
\multirow{2}{*}{\textbf{Method}} & 
\multicolumn{1}{c}{\textbf{Llama-3.2-3B-Instruct}} & 
\multicolumn{1}{c}{\textbf{Qwen2.5-3B-Instruct}} \\
& Wiki PPL & Wiki PPL \\
\hline
\hline
AWQ & 11.75 & 9.10 \\
\hline
CARVQ-4+AWQ & 12.12 & 9.19 \\
CARVQ-3+AWQ & 12.76 & 9.40 \\
CARVQ-2+AWQ & 15.91 & 10.61 \\
\hline
\end{tabular}}
\caption{CARVQ applied on AWQ-quantized models. We evaluate perplexities of Llama-3.2-3B-Instruct and Qwen2.5-3B-Instruct on Wikitext-2.}
\label{tab:awq_compat}
\end{table}


\subsection{Additional Analysis}

\textbf{Ablation of Quantization.} To better understand the effect of RVQ in our method, we swap the RVQ operation with regular scalar quantization and train the Corrective Adaptor with the same configurations. Table \ref{tab:quant_ablation} evaluates the effect of replacing RVQ with a scalar quantization. In both 3B and 8B pre-trained models, RVQ clearly shows an advantage, especially below 2-bit, where INT1 (binary) quantization completely fails (the perplexity goes up to 14528 while with CARVQ-2, it remains 16.84). 

For the comparison to be fair, the group RVQ’s average bitwidth must be equal to the INT-N ($B_{RVQ}=N$). For that, we must say $B_{RVQ}=(pLh2^k+gLk)/gh=N$, where $p$ is fixed to 16 and $L$ fixed to the CARVQ-L. Due to the $gh|nV$ constraint as well as to avoid odd overflows into the next rows when grouping, it is unadvisable to manipulate $g=1024$ and $h=8$ unless we are scaling them by whole numbers. Thus, this leaves $k=4$, the bit-width determining the centroid counts, to be manipulated. We calculate this for LLaMA-3.2-3B. For CARVQ-3 to be equal to INT-2, we would require $k=3.704$, to which it rounds to the original $k=4$. For CARVQ-2 to be equal to INT-1, we require $k=3$, which we report in Table~\ref{tab:quant_ablation}. However, keep in mind that we decrease the number of centroids, which significantly increases the quantization error that has to be regained by the proposed Corrective Adaptor.

\paragraph{Combining with Transformer Quantization.} AWQ allows models to be quantized to 4-bit with minimal loss to quality and coherence. As the Corrective Adaptor only works on the embedding layer, it is inherently orthogonal to existing post-training quantization operations on transformer layers, assuming such methods do not amplify error cascades. We evaluate the perplexity of W4A16 AWQ \cite{lin2023awq} quantized models with CARVQ. Table \ref{tab:awq_compat} details the result of our comparison. The perplexity loss of CARVQ remains low ($<5$) even on AWQ-quantized models, with CARVQ-3 keeping the loss less than $1.1$ in all tests. Notably, the perplexity loss with CARVQ-4 and CARVQ-3 on Qwen2.5 architecture is kept below $0.3$, showing significant bitwidth reduction with near-lossless performance. Further, as W4A16 AWQ natively uses 4-bit and 16-bit datatypes, we can set $\kappa=4$ and $p=16$ for the proposed CARVQ to be compatible with the machines running such models.

\section{Conclusion}

In this work, we introduced CARVQ, a novel post-training method for LLM embedding layer compression to $\sim$1.6 bits without requiring specialized hardware to support lower-bit storage. By carefully reorganizing the matrix into vector-quantized tables, CARVQ reduces the embedding layer precision without utilizing equally low-bit datatypes. We focused on evaluating varying-complexity models on a diverse set of tasks, showing that CARVQ can achieve lower average bitwidth-per-parameter while maintaining reasonable perplexity and accuracy compared to scalar quantization. Moreover, the CARVQ system can be seamlessly integrated into any hardware supporting 4-bit memory to reduce model memory footprint in significantly memory-constrained devices where every MB counts. CARVQ's compatibility with state-of-the-art transformer quantization methods, which can be highly prone to error propagation, unveils promising potential for future scalability.



\section{Limitations}

Despite its conceptual simplicity, CARVQ cannot be directly applied to transformer layers due to a substantial increase in computational complexity without a specialized lookup-table implementation. Such an implementation is feasible for embedding layers, which operate on discrete token indices, but not for continuous transformer activations. This constraint limits CARVQ’s applicability in larger language models, where the proportion of the embedding layer to the overall model size becomes relatively small, even with compression applied to transformer layers. Furthermore, the Corrective Adaptor in CARVQ does not preserve the structural properties of the original embedding matrix, effectively acting as a coarse additional RVQ iteration. While we did not observe notable degradation in our experiments, this simplification may result in the loss of fine-grained semantic information, potentially affecting performance on tasks that depend on subtle representational nuances. Lastly, we observe that larger models tend to be more tolerant to quantization errors in the embedding layer; however, if the accompanying transformer layer compression is excessively lossy, it may fail to compensate for even small embedding-level errors, diminishing the overall robustness of the model.

%

\bibliography{main}


\end{document}